\documentclass[a4paper, 10pt, conference]{ieeeconf}      
\usepackage{FG2025}
\usepackage{comment}
\usepackage{soul}
\usepackage{hyperref}
\FGfinalcopy 

\IEEEoverridecommandlockouts                              
\def\FGPaperID{****} 

\title{\LARGE \bf
FG 2025 TrustFAA Workshop
}

\author{\parbox{16cm}{\centering
     {\large Jiaee Cheong$^1$, Yang Liu$^2$, Harold Soh$^3$ and Hatice Gunes$^1$}\\
     {\normalsize
     $^1$University of Cambridge, $^2$University of Oulu, $^3$National University of Singapore\\
     }}
}

\begin{document}

\maketitle

\begin{abstract}

With the increasing prevalence and deployment of EmotionAI-powered facial affect analysis (FAA) tools, concerns about the trustworthiness of these systems have become more prominent. 
This first workshop on ``\textit{Towards Trustworthy Facial Affect Analysis: Advancing Insights of Fairness, Explainability, and Safety (TrustFAA)}" aims to bring together researchers who are investigating different challenges in relation to trustworthiness — such as interpretability, uncertainty, biases, and privacy — across various facial affect analysis tasks, including macro-/micro-expression recognition, facial action unit detection, other corresponding applications such as pain and depression detection, as well as human-robot interaction and collaboration.
%
%
%
In alignment with FG 2025's emphasis on ethics, as demonstrated by the inclusion of an Ethical Impact Statement requirement for this year's submissions, this workshop supports FG 2025’s efforts by encouraging research, discussion and dialogue on trustworthy FAA.


\end{abstract}

\section{INTRODUCTION}

\subsection{Motivation:}
With the increasing prevalence and deployment of EmotionAI (EAI)-powered facial affect analysis (FAA) tools, concerns about the trustworthiness of these systems have become more prominent \cite{cheong2021hitchhiker}.
In addition, they are increasingly being used in settings which are likely to have direct and profound impact on human lives 
ranging from autonomous driving systems \cite{ngxande2020bias},
education \cite{mejbri2022trends} and healthcare settings \cite{sangeetha2024empirical,cheong2023towards}.
EAI applications often introduce unique real-world challenges which are currently under-investigated within existing trustworthy machine learning (ML) literature.

Trustworthiness can be evaluated across various dimensions \cite{liu2022trustworthy}. 
This first workshop on ``\textit{Towards Trustworthy Facial Affect Analysis: Advancing Insights of Fairness, Explainability, and Safety (TrustFAA)}" aims to bring together researchers who are investigating different challenges in relation to trustworthiness, such as 
interpretability \cite{cambria2024senticnet}, 
explainability \cite{cortinas2023toward}
uncertainty \cite{liu2023uncertain,kuzucu2024uncertainty,cetinkaya2024ranked}, 
biases \cite{xu2020investigating,cheong2023s}, and 
privacy, across various facial affect analysis tasks, including macro-/micro-expression recognition, facial expression recognition \cite{cheong2023counterfactual,Cheong_2023_WACV}, facial action unit detection \cite{liu2022uncertain,churamani2022domain}, as well as other corresponding facial affect analysis such as pain or depression detection \cite{cheong2023towards,cheong_Ufair,cheong2024fairrefuse,kwok2025machine}.

\subsection{Expected outcomes}
The main objective of this workshop is to bring together a multidisciplinary group of researchers to identify and address key challenges to encourage discussion and explore new methodologies that promote the trustworthiness of EAI in the context of FAA tasks.
This workshop aims at: 
\begin{enumerate}
    \item investigating how to incorporate advances in trustworthy machine learning into FAA-related tasks; 
    \item advancing research on trustworthy ML for FAA by understanding the existing gaps as well as open challenges; 
    \item promoting further inclusion on adjacent topics such as ethics and privacy for FAA-related use-cases. 

\end{enumerate}

The workshop will facilitate the exchange of ideas through paper solicitation with oral and poster presentations as well as invited talks and participatory discussions. 
Accepted papers will published as part of the FG workshop proceedings. 

\subsection{Impact}
Promoting the advancement of trustworthy facial affect analysis can have a wide impact across diverse fields such as 
healthcare
\cite{sangeetha2024empirical},
education \cite{mejbri2022trends},
affective computing \cite{cambria2024senticnet,cortinas2023toward}, 
social robots deployment and human-robot interaction (HRI) \cite{cheong2023s, spitale2024hri,cheong2024small} and collaboration \cite{kok2020trust}.
Given the innately high-stakes and sensitive nature of the above use cases, it is essential that researchers ensure the \textit{trustworthiness} of these systems and
adopt measures aligned with ethical guidelines \cite{european2019ethics,devillers2023ethical,saxena2024ethical}.
We anticipate that the talks, presentations, research contributions and discussions will assist researchers in  affective computing, trustworthy ML and HRI to develop trustworthy Emotion AI for all.
%

\section{WORKSHOP OVERVIEW}
TrustFAA is a half-day, hybrid workshop focused on exploring topics related to the trustworthiness of EAI-powered FAA, especially focusing on concerns of fairness, explainability, and safety. 
The proposed workshop includes:

\vspace{0.2cm}
\noindent\textbf{Invited Talks:} The list of potential invited speakers are as follows:
    \begin{itemize}
        \item Professor Brandon Booth.
        \item Professor Akane Sano.
        \item Professor Ali Etemad.
    \end{itemize}  

\vspace{0.2cm}
\noindent\textbf{Oral and Poster Presentations:} The authors of accepted research papers will be invited to present their work as a 8-minute oral presentation which will be followed by a 2-minute Q\&A session.

\noindent\textbf{Tentative schedule:} 
\begin{itemize}
    \item \textbf{9:00-9:10 AM:} Welcome and opening remarks.
    \item \textbf{9:10-9:40 AM:} Invited Speaker \#1.
    \item \textbf{9:40-10:00 AM:} Invited Speaker \#2.
    \item \textbf{10:00-11:30 AM:} Oral session for accepted papers.
    \item \textbf{11:30-12:00 AM:} Invited Speaker \#3.
    \item \textbf{12:00-12:30 PM:} Poster session and discussion.
    \item \textbf{12:30 PM:} Closing remarks.
\end{itemize}

\subsection{Scope List}

We are interested in papers that relate to 
trustworthy ML for affect analysis. The following topics would be of interest. However, the list is intended to be illustrative, not exhaustive:
%
\begin{itemize}
    \item [-] Trustworthy ML/AI methods for FAA, incl. macro-/micro-express recognition, action unit detection, valence \& arousal estimation, etc.
    \item [-] Fairness and bias mitigation in FAA, incl. cross-cultural emotion analysis, reducing gender and racial biases, and assessing equity, etc.
    \item [-] Robustness and uncertainty under real-world variability, incl. trustworthiness in dynamic environments, and adaption to distributional shifts over time, etc.
    \item [-] User-centered explainability in sensitive domains, with a focus on usability for end-users, intuitive interfaces and explanation methods tailored to non-experts, and decision reliability.
    \item [-] Privacy-preserving FAA for sensitive data applications, incl. de-identification technologies,  federated learning, and secure computation.
    \item [-] Assessment and standardization of trustworthy FAA metrics, such as benchmarks and evaluation protocols. 
    \item [-] Ethical and social impacts, incl. data collection guidelines, data transparency, and well-being influences, etc.
\end{itemize}

\subsection{Paper Submission and Review Procedure} 

We invite authors to submit their contributions as 
either a regular paper (6-8 pages long) or position papers (2-4 pages long), using the format provided by FG2025: 
\href{https://fg2025.ieee-biometrics.org/participate/call-for-papers/}
{here}.
The papers should highlight the methodological novelty, experimental results, technical reports and case studies focused on TrustFAA. 
%
%
All submissions will be peer-reviewed for their novelty, relevance, contribution to the field, and technical soundness. 
Authors are also given an option to provide an optional ethical considerations statement and adverse impact statement which will not count towards their total page limit.

The paper submission process will be handled via Easychair.
%
The papers will be reviewed via a double-blind reviewing process by the programme committee. We will solicit both internal and external reviewers from our existing research networks.
We estimate to receive approximately 10-15 submissions with no specific target acceptance rate. Papers will be accepted based on its quality and relevance to the workshop. Accepted papers will published as part of the FG workshop proceedings.
The tentative paper submission and review schedule is as follows: 
\begin{itemize}
    \item Paper submission: \textbf{10 March 2025}
    \item Notification to authors: \textbf{30 March 2025}
    \item Camera ready: \textbf{21 April 2025}
\end{itemize}


\subsection{Planned Advertisement}

The workshop will be advertised to various members of the FG community using the different mailing lists.
%
Additionally, the workshop will also be advertised to 
other adjacent research communities such as the Human-Robot Interaction (HRI) and Affective Computing (AC) communities.
The workshop will also announced using a dedicated website: \href{https://sites.google.com/view/fg2025-trustfaa/home}{FG2025-TrustFAA}.
%
%
The workshop will also be advertised on social media channels such as LinkedIN and X (previously Twitter). The workshop is expected to garner the attention of around 45-50 attendees from the Trustworthy ML, facial affect analysis and affective computing research communities.

\section{Organizers}

\noindent\textbf{Jiaee Cheong} is a doctoral student at the University of Cambridge and the Alan Turing Institute, United Kingdom.
Her research has been funded by the Alan Turing studentship, the Cambridge Trust and the Leverhulme Trust.
Her research interests lie at the intersection of trustworthy and human-centred ML for affective computing and healthcare.

\textit{Relevant publications:}
\begin{enumerate}
\item \textbf{J. Cheong}, S. Kalkan and H. Gunes. FairReFuse: Referee-Guided Fusion for
Multimodal Causal Fairness in Depression Detection. IJCAI-AI for Good, 2024.
\item \textbf{J. Cheong*}, Micol Spitale* and H. Gunes “It’s not Fair!” – Fairness for a Small
Dataset of Multi-Modal Dyadic Mental Well-being Coaching. ACII, 2023. 
\item \textbf{J. Cheong}, S. Kuzucu, S. Kalkan and H. Gunes. Towards Gender Fairness for
Mental Health Predictions. IJCAI-AI for Good, 2023.
\end{enumerate}

\vspace{0.2cm}
\noindent\textbf{Yang Liu} is currently a postdoctoral research fellow with the Center for Machine Vision and Signal Analysis, University of Oulu, Finland, and a visiting scholar with the Department of Computer Science, University of Cambridge, United Kingdom. He was a researcher with Haaga-Helia University of Applied Science, Finland, and a visiting scholar with Hong Kong Baptist University, in 2023. Dr. Liu is the PI of the North Ostrobothnia Regional Fund and the Post-doc Grant, in 2023 and 2024 respectively, carrying out research on digital pain detection through reliable affective computing methods. He has authored and co-authored over 25 published papers mainly focusing on affective intelligence. Dr. Liu has served as Guest Associate Editor of Frontiers in Psychology and Frontiers in Human Neurosciences, and reviewers for prestigious journals (e.g, TAFFC, TMM, TCSVT, TIFS, TSC, IOTJ, PR) and conferences (e.g., ECCV, IJCAI, FG, ICASSP, ICIP, and ICPR). He is a member of FCAI and CAAI, and was a tutorial organizer of HHAI 2024 (Malmö, Sweden). His research interests include multimodal affective computing, trustworthy ML, and AI for medicine.

\textit{Relevant publications:}
\begin{enumerate}
    \item Tutuianu, G. I.$^\dagger$, \textbf{Liu, Y.}$^{\dagger, \ast}$, Alamäki, A., \& Kauttonen, J. Benchmarking deep facial expression recognition: An extensive protocol with balanced dataset in the wild. Engineering Applications of Artificial Intelligence, 2024.
    \item \textbf{Liu, Y.}, Zhang, X., Kauttonen, J., \& Zhao, G. Uncertain facial expression recognition via multi-task assisted correction. IEEE Tran. on Multimedia, 2024.
    \item \textbf{Liu, Y.}, Zhang, X., Li, Y., Zhou, J., Li, X., \& Zhao, G. Graph-based facial affect analysis: A review. IEEE Tran. on Affective Computing, 2023. .
\end{enumerate}

\vspace{0.2cm}
\noindent\textbf{Harold Soh}
is an Assistant Professor of Computer Science at the National University of Singapore, where he leads the Collaborative Learning and Adaptive Robots (CLeAR) group. He completed his Ph.D. at Imperial College London, focusing on online learning for assistive robots.
His research primarily involves machine learning, particularly generative AI, and decision-making in trustworthy collaborative robots. His contributions have been recognized with a R:SS Early Career Spotlight in 2023, best paper awards at IROS’21 and T-AFFC’21, and several nominations (R:SS’18, HRI’18, RecSys’18, IROS’12). Harold has played significant roles in the HRI community, most recently as co-Program Chair of ACM/IEEE HRI’24. He is an Associate Editor for the ACM Transactions on Human Robot Interaction, Robotics Automation and Letters (RA-L), and the International Journal on Robotics Research (IJRR). He is a Principal Investigator at the Smart Systems Institute and a co-founder of TacnIQ, a startup developing touch-enabled intelligence.

\textit{Relevant publications:}
\begin{enumerate}
\item X. Shen, H. Brown, J. Tao, M. Strobel, Y. Tong, A. Narayan, \textbf{H. Soh},
and F. Doshi-Velez. Directions of technical innovation for regulatable
AI systems. Commun. ACM, 67(11):82–89, Oct. 2024.
\item S. Balakrishnan, J. Bi, and \textbf{H. Soh}. SCALES: From fairness principles to
constrained decision-making. In Proceedings of the 2022 AAAI/ACM
Conference on AI, Ethics, and Society, pages 46–55, 2022.
\item B. C. Kok and \textbf{H. Soh}. Trust in robots: Challenges and opportunities.
Current Robotics Reports, 1(4): 297–309, 2020.
\end{enumerate}

\vspace{0.2cm}
\noindent\textbf{Hatice Gunes}
is an internationally recognized scholar and a Full Professor of Affective Intelligence and Robotics at the University of Cambridge, UK. She is a former President of the Association for the Advancement of Affective Computing (AAAC) and was a Faculty Fellow of the Alan Turing Institute – UK’s national centre for data science and artificial intelligence. 
Prof Gunes spearheads award-winning research on multimodal, social, and affective intelligence for AI systems, particularly embodied agents and robots, by cross-fertilizing research in the fields of Machine Learning, Affective Computing, Social Signal Processing, and Human Nonverbal Behaviour Understanding.

\textit{Relevant publications:}
\begin{enumerate}
\item M. Axelsson, M. Spitale, and \textbf{H. Gunes}. Robots as mental well-being
coaches: Design and ethical recommendations. ACM Tran. on
 HRI, 2024.

\item I. Hupont, S. Tolan, \textbf{H. Gunes}, and E. Gomez. The landscape of facial
processing applications in the context of the european ai act and the
development of trustworthy systems. Nature Scientific Reports, 2022.

\item T. Xu, J. White, S. Kalkan, and \textbf{H. Gunes}.  Investigating Bias
and Fairness in Facial Expression Recognition. Proc. ECCV 2020
Workshops, 2020.

\end{enumerate}

\bibliographystyle{ieee}
\bibliography{egbib}

\begin{thebibliography}{10}\itemsep=-1pt

\bibitem{cambria2024senticnet}
E.~Cambria, X.~Zhang, R.~Mao, M.~Chen, and K.~Kwok.
\newblock Senticnet 8: Fusing emotion ai and commonsense ai for interpretable, trustworthy, and explainable affective computing.
\newblock In {\em HCII}, 2024.

\bibitem{cetinkaya2024ranked}
B.~Cetinkaya, S.~Kalkan, and E.~Akbas.
\newblock Ranked: Addressing imbalance and uncertainty in edge detection using ranking-based losses.
\newblock In {\em Proceedings of the IEEE/CVF Conference on Computer Vision and Pattern Recognition}, pages 3239--3249, 2024.

\bibitem{cheong_Ufair}
J.~Cheong, A.~Bangar, S.~Kalkan, and H.~Gunes.
\newblock U-fair: Uncertainty-based multimodal multitask learning for fairer depression detection.
\newblock 2024.

\bibitem{cheong2024fairrefuse}
J.~Cheong, S.~Kalkan, and H.~Gunes.
\newblock Fairrefuse: Referee-guided fusion for multimodal causal fairness in depression detection.

\bibitem{cheong2021hitchhiker}
J.~Cheong, S.~Kalkan, and H.~Gunes.
\newblock The hitchhiker’s guide to bias and fairness in facial affective signal processing: Overview and techniques.
\newblock {\em IEEE Signal Processing Magazine}, 38(6):39--49, 2021.

\bibitem{Cheong_2023_WACV}
J.~Cheong, S.~Kalkan, and H.~Gunes.
\newblock Causal structure learning of bias for fair affect recognition.
\newblock In {\em WACV 2023}, pages 340--349, Jan 2023.

\bibitem{cheong2023counterfactual}
J.~Cheong, S.~Kalkan, and H.~Gunes.
\newblock Counterfactual fairness for facial expression recognition.
\newblock In {\em ECCV 2022 Workshops}, pages 245--261. Springer, 2023.

\bibitem{cheong2023towards}
J.~Cheong, S.~Kuzucu, S.~Kalkan, and H.~Gunes.
\newblock Towards gender fairness for mental health prediction.
\newblock In {\em IJCAI}, pages 5932--5940, 2023.

\bibitem{cheong2023s}
J.~Cheong, M.~Spitale, and H.~Gunes.
\newblock “it’s not fair!”--fairness for a small dataset of multi-modal dyadic mental well-being coaching.
\newblock In {\em 2023 11th International Conference on Affective Computing and Intelligent Interaction (ACII)}, pages 1--8. IEEE, 2023.

\bibitem{cheong2024small}
J.~Cheong, M.~Spitale, and H.~Gunes.
\newblock Small but fair! fairness for multimodal human-human and robot-human mental wellbeing coaching.
\newblock {\em arXiv preprint arXiv:2407.01562}, 2024.

\bibitem{churamani2022domain}
N.~Churamani, O.~Kara, and H.~Gunes.
\newblock Domain-incremental continual learning for mitigating bias in facial expression and action unit recognition.
\newblock {\em IEEE Transactions on Affective Computing}, 14(4):3191--3206, 2022.

\bibitem{cortinas2023toward}
K.~Corti{\~n}as-Lorenzo and G.~Lacey.
\newblock Toward explainable affective computing: A review.
\newblock {\em IEEE Transactions on Neural Networks and Learning Systems}, 2023.

\bibitem{devillers2023ethical}
L.~Devillers and R.~Cowie.
\newblock Ethical considerations on affective computing: an overview.
\newblock {\em Proceedings of the IEEE}, 2023.

\bibitem{european2019ethics}
S.~European~Commission et~al.
\newblock Ethics guidelines for trustworthy ai.
\newblock {\em Publications Office}, 2019.

\bibitem{kok2020trust}
B.~C. Kok and H.~Soh.
\newblock Trust in robots: Challenges and opportunities.
\newblock {\em Current Robotics Reports}, 1(4):297--309, 2020.

\bibitem{kuzucu2024uncertainty}
S.~Kuzucu, J.~Cheong, H.~Gunes, and S.~Kalkan.
\newblock Uncertainty as a fairness measure.
\newblock {\em Journal of Artificial Intelligence Research}, 81:307--335, 2024.

\bibitem{kwok2025machine}
A.~M.~H. Kwok, J.~Cheong, S.~Kalkan, and H.~Gunes.
\newblock Machine learning fairness for depression detection using eeg data.
\newblock {\em arXiv preprint arXiv:2501.18192}, 2025.

\bibitem{liu2022trustworthy}
H.~Liu, Y.~Wang, W.~Fan, X.~Liu, Y.~Li, S.~Jain, Y.~Liu, A.~Jain, and J.~Tang.
\newblock Trustworthy ai: A computational perspective.
\newblock {\em ACM Transactions on Intelligent Systems and Technology}, 14(1):1--59, 2022.

\bibitem{liu2022uncertain}
Y.~Liu, X.~Zhang, J.~Kauttonen, and G.~Zhao.
\newblock Uncertain label correction via auxiliary action unit graphs for facial expression recognition.
\newblock In {\em 2022 26th International Conference on Pattern Recognition (ICPR)}, pages 777--783. IEEE, 2022.

\bibitem{liu2023uncertain}
Y.~Liu, X.~Zhang, J.~Kauttonen, and G.~Zhao.
\newblock Uncertain facial expression recognition via multi-task assisted correction.
\newblock {\em IEEE Transactions on Multimedia}, 2023.

\bibitem{mejbri2022trends}
N.~Mejbri, F.~Essalmi, M.~Jemni, and B.~A. Alyoubi.
\newblock Trends in the use of affective computing in e-learning environments.
\newblock {\em Education and Information Technologies}, pages 1--23, 2022.

\bibitem{ngxande2020bias}
M.~Ngxande, J.-R. Tapamo, and M.~Burke.
\newblock Bias remediation in driver drowsiness detection systems using generative adversarial networks.
\newblock {\em IEEE Access}, 8:55592--55601, 2020.

\bibitem{sangeetha2024empirical}
S.~Sangeetha, R.~R. Immanuel, S.~K. Mathivanan, J.~Cho, and S.~V. Easwaramoorthy.
\newblock An empirical analysis of multimodal affective computing approaches for advancing emotional intelligence in artificial intelligence for healthcare.
\newblock {\em IEEE Access}, 2024.

\bibitem{saxena2024ethical}
C.~Saxena.
\newblock Ethical considerations in affective computing.
\newblock In {\em Affective Computing for Social Good: Enhancing Well-being, Empathy, and Equity}, pages 241--251. Springer, 2024.

\bibitem{spitale2024hri}
M.~Spitale, R.~Stower, M.~T. Parreira, E.~Yadollahi, I.~Leite, and H.~Gunes.
\newblock Hri wasn’t built in a day: A call to action for responsible hri research.
\newblock In {\em 2024 33rd ROMAN}, pages 696--702. IEEE, 2024.

\bibitem{xu2020investigating}
T.~Xu, J.~White, S.~Kalkan, and H.~Gunes.
\newblock Investigating bias and fairness in facial expression recognition.
\newblock In {\em Computer Vision--ECCV 2020 Workshops: Glasgow, UK, August 23--28, 2020, Proceedings, Part VI 16}, pages 506--523. Springer, 2020.

\end{thebibliography}

\end{document}